\documentclass{article}
\usepackage{spconf,amsmath,graphicx}

\usepackage{enumitem}
\setlist{nosep, leftmargin=14pt}

\usepackage{mwe} 

\usepackage{spconf,amsmath,graphicx}
\usepackage{color}
\usepackage{upgreek}
\usepackage{bbm}
\usepackage[utf8]{inputenc}
\usepackage[pagebackref,breaklinks,colorlinks]{hyperref}
\usepackage{tabularx}
\usepackage{multirow}
\usepackage{amsfonts}

\definecolor{cb_orange}{rgb}{1.0,0.51,0.0}
\definecolor{cb_blue}{rgb}{0.22,0.49,0.72}
\definecolor{cb_green}{rgb}{0.3,0.67,0.29}
\definecolor{cb_red}{rgb}{0.89,0.1,0.11}
\definecolor{cb_pink}{rgb}{1, 0, 0.4}

\title{CGAM: Click-Guided Attention Module for Interactive Pathology \\Image Segmentation via Backpropagating Refinement}
\name{Seonghui Min \qquad
  Won-Ki Jeong $^{*}$ \qquad \thanks{*Corresponding author: wkjeong@korea.ac.kr}
}
\address{
Department of Computer Science and Engineering, Korea University,  Seoul, Korea}
\begin{document}
%
\maketitle
\begin{abstract}
%
Tumor region segmentation is an essential task for the quantitative analysis of digital pathology. 
Recently presented deep neural networks have shown state-of-the-art performance in various image-segmentation tasks. 
However, because of the unclear boundary between the cancerous and normal regions in pathology images, despite using modern methods, it is difficult to produce satisfactory segmentation results in terms of the reliability and accuracy required for medical data. 
In this study, we propose an interactive segmentation method that allows users to refine the output of deep neural networks through click-type user interactions. 
The primary method is to formulate interactive segmentation as an optimization problem that leverages both user-provided click constraints and semantic information in a feature map using a click-guided attention module (CGAM).
Unlike other existing methods, CGAM avoids excessive changes in segmentation results, which can lead to the overfitting of user clicks. 
Another advantage of CGAM is that the model size is independent of input image size. 
Experimental results on pathology image datasets indicated that our method performs better than existing state-of-the-art methods.
\end{abstract}
%

\section[Introduction]
{Introduction\\}
\label{sec:intro}
Segmenting tumor area in whole-slide images (WSI) is an important task in digital pathology, as it serves as a basis for the diagnosis of a target lesion.
However, the difference in visual features, including the color and texture of malignant and normal regions, is insignificant in particular histopathological images. 
Because of this innate biological property, even experts in this domain need considerable time to accurately distinguish these areas with the naked eye. 
In addition, it is difficult to capture precise boundaries for classifying malignant regions using conventional automated segmentation methods that mostly rely on edges. 
To this end, interactive segmentation modifies automated segmentation methods to enable user interactions~\cite{9434105}. 
This allows users to quickly obtain high-quality segmentation results by providing interactions that reflect their intentions.\par

Commonly used user-interaction types include bounding boxes~\cite{wu2014milcut, rother2004grabcut}, scribbles~\cite{grady2006random, gulshan2010geodesic}, and clicks~\cite{jang2019interactive, sofiiuk2020f, lin2022generalizing, xu2016deep}. 
Among these, we specifically focused on 
click-based interactive segmentation in which users provide positive/negative clicks to differentiate between foreground and background regions. 
Prior to deep learning, conventional approaches \cite{rother2004grabcut, grady2006random, gulshan2010geodesic} considered interactive segmentation as an optimization problem. 
Because semantic information has not been fully exploited with many built-in heuristics, these methods require large amounts of user interactions. 
%
Deep learning-based interactive segmentation methods~\cite{xu2016deep, maninis2018deep} improve the segmentation accuracy of deep neural networks~\cite{long2015fully, he2016deep} by incorporating user interactions. 
%
While showing outstanding performance compared to conventional methods, existing deep learning-based interactive segmentation methods rely
heavily on 
high-level semantic priors and perform poorly for object classes not seen during training. 
%
%
%
%
%
%
%

\begin{figure*}[ht!]
\centering\includegraphics[width=0.9\textwidth]{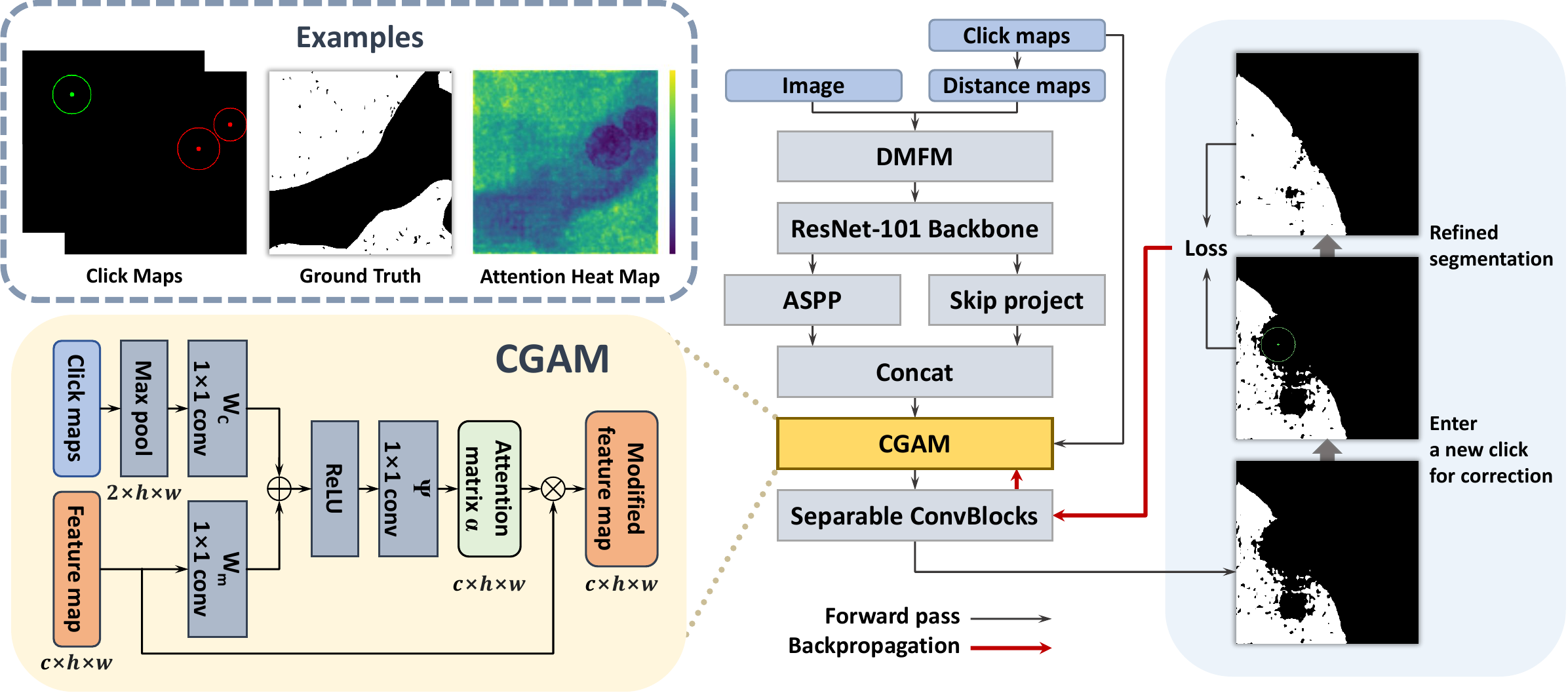}
\caption{Overview of the proposed method, and the structure of CGAM. The user inputs a click where correction is required in the segmentation result. CGAM takes click maps and an intermediate feature map as input. It weights the feature map to highlight the target region of segmentation. By backpropagating the loss computed from user-provided clicks and output prediction, the parameters of CGAM are updated to produce an improved segmentation result. This process is repeated each time the user provides a new click. Red and green represent positive and negative clicks, respectively. The attention radius is visualized with a circle.
}
\label{fig:overview}
\end{figure*}
%
%
%
%
%
%
%
%

Recently, 
backpropagating refinement scheme (BRS) \cite{jang2019interactive} addressed this issue by integrating optimization-based and deep learning-based methods.
BRS sets the interaction maps entered into the network as trainable parameters.
By backpropagating the loss calculated by prediction and user clicks, BRS fine-tunes the interaction maps in an online manner.
Feature backpropagating refinement scheme ($f$-BRS)~\cite{sofiiuk2020f} is an improvement to the previous method in terms of inference time and computational budget by inserting a set of auxiliary parameters after the intermediate network layer for optimization.
Backpropagation only through a subpart of the network improves the efficiency of $f$-BRS.
%
In a follow-up study, Lin \textit{et al.}~\cite{lin2022generalizing} proposed generalized backpropagating refinement scheme (G-BRS), advanced layer architectures that enable more delicate refinement. 
However, the above methods optimize additional modules only through the minimization of loss calculated by limited user interactions, 
causing unwanted overall changes owing to overfitting.
%
%
\par
In this study, we propose a new click-guided attention module (\textit{CGAM}) 
for BRS-based interactive segmentation. 
%
CGAM addresses the overfitting issue of existing BRS-based methods 
by directly receiving click maps and a feature map from where it is inserted as a module input and utilizing them for optimization. 
CGAM enforces the desired specification on the result of a deep learning model by restricting the feature space with self-attention and the additional guidance of click maps.
Furthermore, in contrast to  G-BRS, 
the model size of CGAM is independent of the input image size, allowing the method to easily handle large-scale images.
%
%
We demonstrate the segmentation performance of CGAM over existing deep learning-based interactive segmentation refinement methods on the PAIP2019 challenge dataset~\cite{kim2021paip}.

\section{METHOD}

\label{sec:format}
\subsection{Architecture Overview}
An overview of the proposed method is shown in Fig.~\ref{fig:overview}. 
%
To compare the proposed model, $f$-BRS, 
and G-BRS directly, 
we chose the standard DeepLabV3+ with ResNet-101 containing a distance maps fusion module (DMFM) proposed in \cite{sofiiuk2020f} as the basis architecture. 
As in~\cite{sofiiuk2020f, lin2022generalizing}, 
CGAM is inserted after the atrous spatial pyramid pooling (ASPP) layer in the DeepLabV3+ decoder.
CGAM modifies the feature map of the corresponding location by receiving the feature map and click maps as inputs. 
The output logits are generated as the modified feature map passes through the rest of the network. 
In this study, we define interactive segmentation as an optimization problem and solve it with respect to the parameters of CGAM. 
Thus, the optimization loss is computed from the output logits and user-provided clicks. 
By backpropagating the loss, CGAM is updated for better segmentation performance in an online manner.

\par

Assume $f$ is a function implemented by the basis network.
With input image $I$ and click maps $C$, the intermediate feature map of the location where CGAM is inserted is defined as $g(I,C)$. 
Using $h$ to denote a function that the network head implements, $f$ can be represented as $f(I,C):= h(g(I,C))$. 
We express the whole process $\hat{f}$ with CGAM parameterized by $\theta$ inserted as follows:
\begin{align}
   \hat{f}(I,C;\theta) = h(\theta(g(I,C),\ C)).
\end{align}
We define a set of user-provided clicks as $\{(u_i,v_i,l_i,r_i)\}^N_{i=1}$ where $(u,v)$, $l\in{\{-1,1\}}$, and $r$ represent the coordinates, label, and radius of each click, respectively. $M\in{\{0,1\}}^{H\times W}$ is a binary mask generated using the newest click and selects the region outside of $r$. 
The optimization problem is formulated as a minimization for the following loss similar to that of \cite{lin2022generalizing}:
\begin{multline}
\label{eq:loss}
    \mathcal{L}_t(I,C_t) = \min_{\theta_t}\mathbb{E}_{i\in[1,t]} [max(l_i - \hat{f}(I,C_t;\theta_t)_{u_i,u_v},0)]^2 \\
    + \lambda \|M\odot(\hat{f}(I,C_t; \theta_{t-1})-\hat{f}(I,C_t;\theta_t))\|^2_2,
\end{multline}
where $\odot$ is Hadamard product, and $t\in[1,N]$ is an interaction step. The first term enforces the correct output segmentation corresponding to the user-provided clicks, and the second term prevents excessive modification to avoid overfitting. The scaling constant $\lambda$ regulates the trade-off between the two terms.

\begin{figure*}[t!]
\centering
\includegraphics[width=1.0\textwidth]{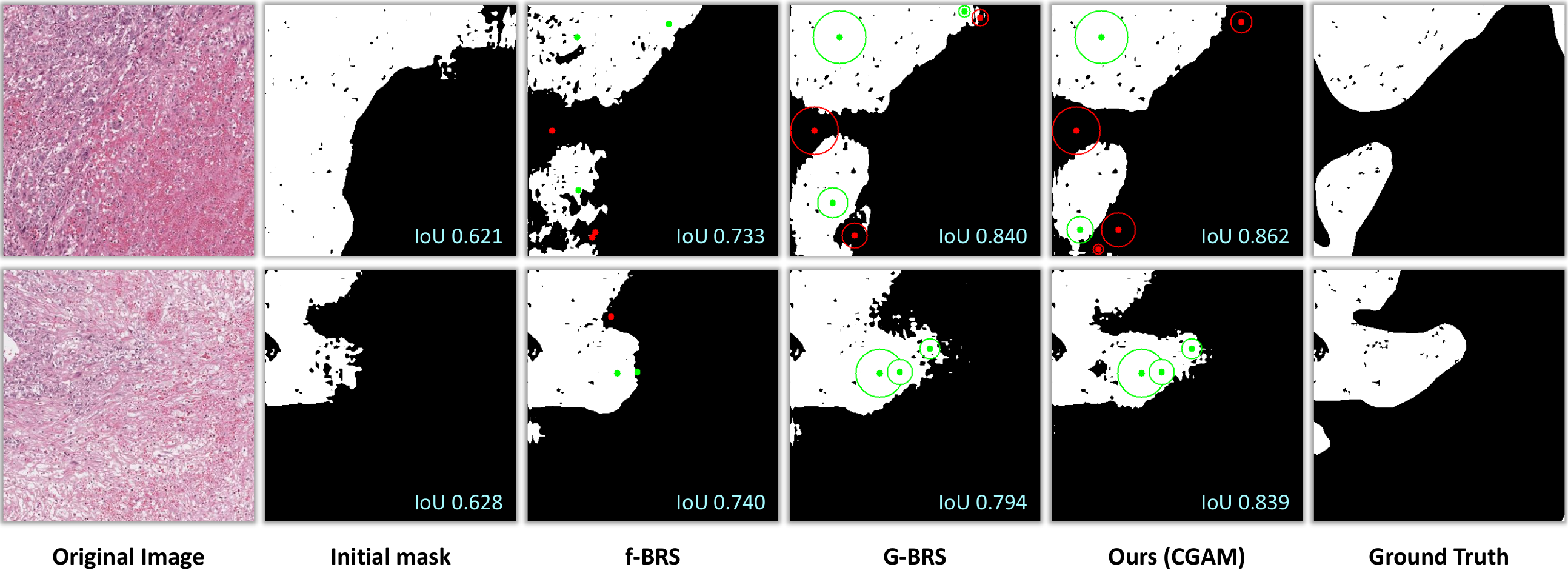}
\caption{Comparison of various BRS-based interactive segmentation results. Each method used the same number of clicks.}
%
\end{figure*}

\subsection{Click-Guided Attention Module}
\label{ssec:subhead}
CGAM is a self-attention module that specializes in interactive segmentation, inspired by self-attention methods \cite{oktay2018attention, wang2018non}. 
By receiving assistance from the additional guidance of click maps, CGAM highlights feature responses in regions reflecting user intention.
Fig.~\ref{fig:overview} left illustrates the pipeline of CGAM.
%
%
%
%
%
%
%
By denoting the input feature map as $g(I,C)=m\in\mathbb{R}^{c\times hw}$, attention matrix $\alpha\in\mathbb{R}^{c\times hw}$ is obtained as follows:
\begin{align}
   \alpha = \ \psi^T (ReLU \ (W_C ^T C_d \ + \  W_m ^T m)),
\end{align}
where $C_d\in\mathbb{R}^{2\times hw}$ represents click maps downsampled to the resolution of $m$.
The linear transformations with weight matrices $W_C\in\mathbb{R}^{2\times {c \over 2}}$, $W_m\in\mathbb{R}^{c\times {c \over 2}}$, and $\psi\in\mathbb{R}^{{c \over 2} \times c}$ are implemented as a 1$\times$1 convolution.
The output of CGAM, a modified feature map $\hat{m}\in\mathbb{R}^{c\times hw}$, is then finally calculated as follows:
\begin{align}
   \hat{m} = m \ \odot \ \alpha.
\end{align}
CGAM preserves the initial behavior of the network before it learns through backpropagation by setting the initial value of $\alpha$ to one.\par
%
%
%
%
%
%
%
%
%
It can be observed from the attention heat map in Fig.~\ref{fig:overview} that CGAM focuses on important regions by exploiting information in the click map.
%
%
%
%
%
%
The attention matrix assigns element-wise weights to the feature map, which enables local refinement.
This operation sets the CGAM free from the dependency of the input image size on G-BRS. 
%

\section{EXPERIMENT}
\label{sec:pagestyle}
\subsection{Data Description}
\label{ssec:subhead}
The whole-slide image (WSI) dataset 
used in our experiment was 
from the PAIP2019 challenge~\cite{kim2021paip}. 
After scaling at 5$\times$ magnification, 
because interactive refinement of segmentation results is mainly required at the boundary of the tumor, patches with tumor areas accounting for 20\% to 80\% of the total area were considered boundary regions and extracted.

\subsection{Implementation Details}
\label{ssec:subhead}
We trained our network on the pathology dataset with 5190 patches using the normalized focal loss proposed in \cite{sofiiuk2019adaptis}. 
We sampled the clicks during training following the standard procedure of \cite{xu2016deep}. 
The maximum number of clicks per image was set as 20, limiting the number of positive and negative clicks to less than 10. 
We used the Adam optimizer with $\beta_1$ = 0.9, $\beta_2$ = 0.999, and trained the networks for 120 epochs. 
We set the learning rate as 5 $\times$ $10^{-4}$ for the first 100 epochs, and 5 $\times$ $10^{-5}$ for the last 20 epochs. \par

For the inference time optimization, we also used the Adam optimizer with $\beta_1$ = 0.9 and $\beta_2$ = 0.999. 
We performed back-propagation for 20 iterations for each click. 
We set the learning rate as 5 $\times$ $10^{-2}$ and $\lambda$ as 1. 
We conducted an experiment on a Windows 10 PC equipped with an NVIDIA RTX 3090 GPU.

\subsection{Evaluation Protocol}
\label{ssec:eval_protocal}
For fair comparisons, we used the automatic click generation strategy proposed in \cite{lin2022generalizing}: The class of the following click was determined based on whether the dominant prediction error type was false positive or false negative. 
The click was placed at the point where the corresponding error region had the maximum Euclidean distance from its boundary. 
The distance was set as the radius of the click. 
This process was repeated until the target Intersection over Union (IoU) or the maximum number of clicks was reached.

\begin{figure}[t]
\centering
\includegraphics[width=0.5\textwidth]{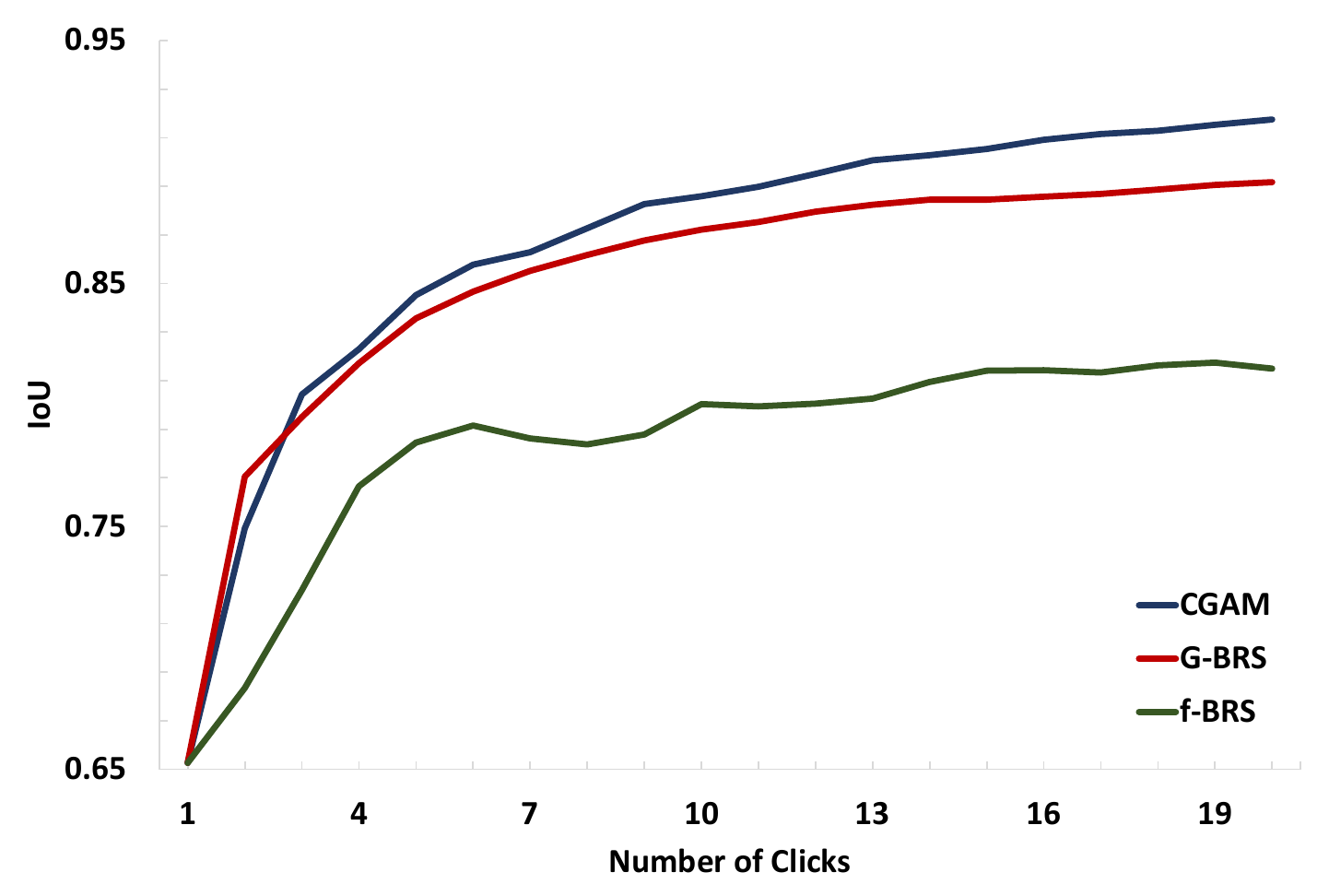}
\caption{Comparison of the IoU scores with respect to the number of clicks added by a user. 
}
\label{fig:graph}
\end{figure}

\begin{table}[htb!]
\centering
\scriptsize
{\caption{Evaluation results of pathology image dataset. The best results are in bold.}
\label{table:1}
\renewcommand{\arraystretch}{1.8}
\setlength{\tabcolsep}{4.8pt}
\begin{tabular}{rcccccc}
\hline
 Ev{Method} & \textbf{NoC@85} & \textbf{NoC@90} & \textbf{NoF@85} & \textbf{NoF@90} & \textbf{SPC (s)} & \textbf{Time (s)} \\ \hline\hline
\textbf{$f$-BRS} & 11.59 & 15.43 & 55 & 89 & 0.053 & 107
 \\ \hline
\textbf{G-BRS} & 7.70 & 12.67 & 23 & 58 & \textbf{0.049} & 81
 \\ \hline
\textbf{CGAM} & \textbf{6.38} & \textbf{11.32} & \textbf{8} & \textbf{36} & 0.052 & \textbf{77}
 \\ \hline
\end{tabular}
}
\end{table}

\subsection{Evaluation Metrics}
\label{ssec:subhead}
We set the target IoU as 85\% and 90\%. We limited the maximum number of clicks to 20. We reported the mean number of clicks (NoC) required to achieve the target IoU. We reported the number of failures (NoF) indicating the number of cases in which the target IoU was not reached with the maximum number of clicks. We reported the second per click (SPC) to measure the response time for each click. Finally, we reported the total time required to process the entire dataset.

\section{Results}
\subsection{Comparison}
\label{ssec:subhead}
We evaluated 131 patches of WSI whose initial predictions had IoU scores between 50\% and 70\%. 
We compared CGAM with $f$-BRS 
and G-BRS, 
state-of-the-art BRS-based methods. 
For G-BRS, we selected the G-BRS-bmconv layer with the best performance reported in~\cite{lin2022generalizing}. \par 

Table~\ref{table:1} presents the average NoC, NoF, SPC, and total time of the three methods for target IoUs of 85\% and 90\%. 
CGAM outperformed the other methods in nearly all metrics. 
The NoC results show that users can obtain satisfactory segmentation masks with less effort using CGAM. 
From the NoF results, as compared to the other methods, it is observed that CGAM successfully reached the target IoU in most cases. 
For speed-related metrics, the SPC of CGAM was slightly slower than that of G-BRS. 
However, CGAM reduced the total time required to reach the target IoU with fewer clicks.



\begin{table}[t!]
\centering
\scriptsize
{\caption{Ablation study of click map guidance. In each method, CGAM takes proper click maps, random click maps, and zero tensor as input.}
\label{table:2}
\renewcommand{\arraystretch}{1.7}
\setlength{\tabcolsep}{4.8pt}
\begin{tabular}{lcccccc}
\hline
 \textbf{Method} & \textbf{NoC@85} & \textbf{NoC@90} & \textbf{NoF@85} & \textbf{NoF@90} & \textbf{SPC (s)} & \textbf{Time (s)} \\ \hline\hline
\textbf{Proper} &  \textbf{7.18} & \textbf{11.82} & \textbf{11} & \textbf{36} & \textbf{0.050} & \textbf{77}
 \\ \hline
\textbf{Random} & 7.31 & 12.43 & 19 & 53 & 0.060 & 95
 \\ \hline
\textbf{Zero} & 7.37 & 12.23 & 15 & 48 & 0.060 & 95
 \\ \hline
\end{tabular}
}
\end{table}


\subsection{Ablation Study}
\label{ssec:subhead}
We conducted an ablation study to assess how click maps contribute to the performance of CGAM. 
We tested the following three scenarios;  
The first scenario received clicks on the appropriate coordinates and classes, generated by an automatic click-generation strategy described in Subsection~\ref{ssec:eval_protocal}.
The second scenario assumed inappropriate (incorrect) clicks by generating clicks on random coordinates and classes. 
The third scenario assumed that no clicks were provided using a zero tensor with the same shape as that of the click maps. 
As shown in Table~\ref{table:2}, CGAM achieved the best performance in all metrics when proper click maps were provided. 
Considering that the results of random clicks are worse than those of zero-tensor case, we can confirm that CGAM leverages the information in the click maps for optimization.


\subsection{Discussion}
CGAM outperformed other methods in terms of both accuracy and time. 
%
In particular, the model size of CGAM is fixed and constant regardless of the input size, unlike G-BRS, which expands linearly in proportion to the height and width of the input image; this decreases the number of parameters from 93k to 84k even for small images with 400 $\times$ 400 pixels.
%
%
These results show the efficiency of CGAM and further demonstrate its potential for extending to multi-scale and large image segmentation tasks (e.g., segmentation of WSI).

\section{CONCLUSION}
\label{sec:typestyle}
In this study, we proposed 
CGAM for interactive image segmentation through back-propagating refinement. 
Exploiting the information in click maps by using it as input, CGAM increases the utility of user-provided clicks in interactive segmentation tasks. 
Experiments showed the improved performance of CGAM in pathology image segmentation as compared to other state-of-the-art methods. 
In future work, we plan to extend the current framework to address entire WSI such that it can be flexibly applied in real-world situations.

\section{Compliance With Ethical Standards}
This research study was conducted retrospectively using human subject data made available in open access by PAIP2019. Ethical approval was not required as confirmed by the license attached with the open access data.

\section{Acknowledgments}
This work is supported by the National Research Foundation of Korea (NRF-2019M3E5D2A01063819, NRF-2021R1A6\\A1A13044830), the Institute for Information \& Communications Technology Planning \& Evaluation (IITP-2023-2020-0-01819), and the Korea Health Industry Development Institute (HI18C0316).

\bibliographystyle{IEEEbib}
\bibliography{main.bib}

\begin{thebibliography}{10}

\bibitem{9434105}
Sungduk Cho, Hyungjoon Jang, Jing~Wei Tan, and Won-Ki Jeong,
\newblock ``{DeepScribble: Interactive Pathology Image Segmentation Using Deep
  Neural Networks with Scribbles},''
\newblock in {\em 2021 IEEE 18th International Symposium on Biomedical Imaging
  (ISBI)}, 2021, pp. 761--765.

\bibitem{wu2014milcut}
Jiajun Wu, Yibiao Zhao, Jun-Yan Zhu, Siwei Luo, and Zhuowen Tu,
\newblock ``{MILCut: A Sweeping Line Multiple Instance Learning Paradigm for
  Interactive Image Segmentation},''
\newblock in {\em Proceedings of the IEEE Conference on Computer Vision and
  Pattern Recognition}, 2014, pp. 256--263.

\bibitem{rother2004grabcut}
Carsten Rother, Vladimir Kolmogorov, and Andrew Blake,
\newblock ``{GrabCut: Interactive foreground extraction using iterated graph
  cuts},''
\newblock {\em ACM Transactions on Graphics (ToG)}, vol. 23, no. 3, pp.
  309--314, 2004.

\bibitem{grady2006random}
Leo Grady,
\newblock ``{Random Walks for Image Segmentation},''
\newblock {\em IEEE transactions on pattern analysis and machine intelligence},
  vol. 28, no. 11, pp. 1768--1783, 2006.

\bibitem{gulshan2010geodesic}
Varun Gulshan, Carsten Rother, Antonio Criminisi, Andrew Blake, and Andrew
  Zisserman,
\newblock ``{Geodesic Star Convexity for Interactive Image Segmentation},''
\newblock in {\em 2010 IEEE Computer Society Conference on Computer Vision and
  Pattern Recognition}. IEEE, 2010, pp. 3129--3136.

\bibitem{jang2019interactive}
Won-Dong Jang and Chang-Su Kim,
\newblock ``{Interactive Image Segmentation via Backpropagating Refinement
  Scheme},''
\newblock in {\em Proceedings of the IEEE Conference on Computer Vision and
  Pattern Recognition}, 2019, pp. 5297--5306.

\bibitem{sofiiuk2020f}
Konstantin Sofiiuk, Ilia Petrov, Olga Barinova, and Anton Konushin,
\newblock ``{f-BRS: Rethinking Backpropagating Refinement for Interactive
  Segmentation},''
\newblock in {\em Proceedings of the IEEE Conference on Computer Vision and
  Pattern Recognition}, 2020, pp. 8623--8632.

\bibitem{lin2022generalizing}
Fanqing Lin, Brian Price, and Tony Martinez,
\newblock ``{Generalizing Interactive Backpropagating Refinement for Dense
  Prediction Networks},''
\newblock in {\em Proceedings of the IEEE Conference on Computer Vision and
  Pattern Recognition}, 2022, pp. 773--782.

\bibitem{xu2016deep}
Ning Xu, Brian Price, Scott Cohen, Jimei Yang, and Thomas~S Huang,
\newblock ``{Deep Interactive Object Selection},''
\newblock in {\em Proceedings of the IEEE Conference on Computer Vision and
  Pattern Recognition}, 2016, pp. 373--381.

\bibitem{maninis2018deep}
Kevis-Kokitsi Maninis, Sergi Caelles, Jordi Pont-Tuset, and Luc Van~Gool,
\newblock ``{Deep Extreme Cut: From Extreme Points to Object Segmentation},''
\newblock in {\em Proceedings of the IEEE Conference on Computer Vision and
  Pattern Recognition}, 2018, pp. 616--625.

\bibitem{long2015fully}
Jonathan Long, Evan Shelhamer, and Trevor Darrell,
\newblock ``{Fully Convolutional Networks for Semantic Segmentation},''
\newblock in {\em Proceedings of the IEEE Conference on Computer Vision and
  Pattern Recognition}, 2015, pp. 3431--3440.

\bibitem{he2016deep}
Kaiming He, Xiangyu Zhang, Shaoqing Ren, and Jian Sun,
\newblock ``{Deep Residual Learning for Image Recognition},''
\newblock in {\em Proceedings of the IEEE Conference on Computer Vision and
  Pattern Recognition}, 2016, pp. 770--778.

\bibitem{kim2021paip}
Yoo~Jung Kim, Hyungjoon Jang, Kyoungbun Lee, Seongkeun Park, Sung-Gyu Min,
  Choyeon Hong, Jeong~Hwan Park, Kanggeun Lee, Jisoo Kim, Wonjae Hong, et~al.,
\newblock ``{PAIP 2019: Liver cancer segmentation challenge},''
\newblock {\em Medical Image Analysis}, vol. 67, pp. 101854, 2021.

\bibitem{oktay2018attention}
Ozan Oktay, Jo~Schlemper, Loic~Le Folgoc, Matthew Lee, Mattias Heinrich,
  Kazunari Misawa, Kensaku Mori, Steven McDonagh, Nils~Y Hammerla, Bernhard
  Kainz, et~al.,
\newblock ``{Attention U-Net: Learning Where to Look for the Pancreas},''
\newblock {\em arXiv preprint arXiv:1804.03999}, 2018.

\bibitem{wang2018non}
Xiaolong Wang, Ross Girshick, Abhinav Gupta, and Kaiming He,
\newblock ``{Non-local Neural Networks},''
\newblock in {\em Proceedings of the IEEE Conference on Computer Vision and
  Pattern Recognition}, 2018, pp. 7794--7803.

\bibitem{sofiiuk2019adaptis}
Konstantin Sofiiuk, Olga Barinova, and Anton Konushin,
\newblock ``{AdaptIS: Adaptive Instance Selection Network},''
\newblock in {\em Proceedings of the IEEE International Conference on Computer
  Vision}, 2019, pp. 7355--7363.

\end{thebibliography}



\end{document}